\pdfoutput=1

\documentclass[11pt]{article}
\usepackage[final]{acl}
\usepackage{times}
\usepackage{latexsym}
\usepackage[T1]{fontenc}
\usepackage[utf8]{inputenc}
\usepackage{microtype}
\usepackage{inconsolata}
\usepackage{graphicx}
\usepackage{hyperref}
\usepackage{makecell}

\usepackage{caption}
\usepackage{subcaption}
\usepackage{multirow}
\usepackage{adjustbox}
\usepackage{booktabs}

\title{Judgment-of-Thought Prompting: A Courtroom-Inspired Framework \\ for Binary Logical Reasoning with Large Language Models}
\author{Sungjune Park, Heehwan Kim, Haehyun Cho \and Daeseon Choi \\
         Soongsil University, Korea \\ \texttt{\{joey25,kh2ss9812\}@soongsil.ac.kr}, \texttt{\{haehyun,sunchoi\}@ssu.ac.kr}}

\date{}

\begin{document}
\maketitle

\begin{abstract}
    This paper proposes a novel prompting approach, Judgment of Thought (JoT), specifically tailored for binary logical reasoning tasks. 
Despite advances in prompt engineering, existing approaches still face limitations in handling complex logical reasoning tasks.

To address these issues, JoT introduces a multi-agent approach with three specialized roles---lawyer, prosecutor, and judge---where a high-level model acts as the judge, and lower-level models serve as lawyer and prosecutor to systematically debate and evaluate arguments. 
Experimental evaluations on benchmarks such as BigBenchHard and Winogrande demonstrate JoT's superior performance compared to existing prompting approaches, achieving notable improvements, including 98\% accuracy in Boolean expressions. 
Also, our ablation studies validate the critical contribution of each role, iterative refinement loops, and feedback mechanisms.

Consequently, JoT significantly enhances accuracy, reliability, and consistency in binary reasoning tasks and shows potential for practical applications.





\end{abstract}

\begin{figure*}[t]
\centering
\includegraphics[width=0.95\textwidth]{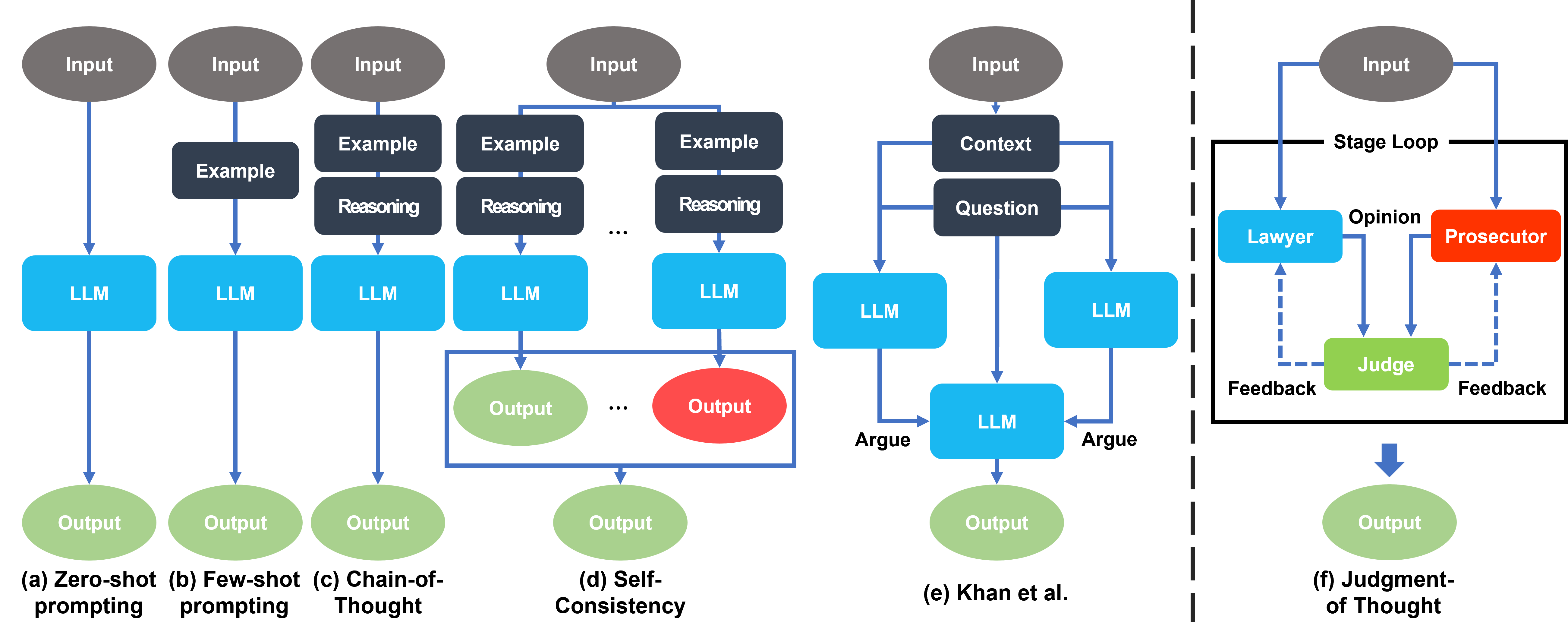} 
\caption{Comparison of Judgment of Thought (ours) with recent prompting strategies.}
\label{fig1}
\end{figure*}

\section{Introduction}
\label{s:intro}
Recent advances in AI and natural language processing (NLP) have brought major changes to many industries~\cite{vaswani2017attention,peters-etal-2018-deep,devlin-etal-2019-bert}. 
In particular, Large Language Models (LLMs) have shown impressive performance on a wide range of language tasks, such as text generation, translation, and sentiment analysis~\cite{floridi2020gpt,touvron2023llama,zhao2023survey,chang2024survey,achiam2023gpt}. 
These models are trained on massive datasets and have learned to understand and generate language in flexible, general-purpose ways\cite{zhang-etal-2024-comprehensive-survey}.

However, to get high-quality results from LLMs, it's important to carefully design the input text---known as a prompt\cite{wahle-etal-2024-paraphrase}. 
The way a prompt is written can greatly affect how accurate, helpful, or logical the model’s output is~\cite{benedetto2024using}. 
This practice, called \textit{prompt engineering}, helps guide LLMs to produce responses that match the user’s goals~\cite{schulhoff2024prompt,sahoo2024systematic,wang2024investigating}.

Many prompting approaches have been proposed to improve reasoning quality. 
These include zero-shot and few-shot prompting, and \textit{Chain-of-Thought (CoT)} prompting, which encourages the model to explain its reasoning step by step~\cite{wei2021finetuned,kaplan2020scaling,touvron2023llama,wei2022chain,wang2022self}. 
More recently, \citet{khan2024debating} showed that debate-style prompting, where a stronger model argues against a weaker one, can lead to more accurate answers—especially when evaluated by a third model acting as a judge.
Despite these advances, current prompting methods still have limitations: especially for binary decisions that require careful reasoning. 
Tasks involving subtle logic, ambiguity, or conflicting claims often lead to inconsistent or incorrect answers. 
Existing methods do not always handle disagreements well or allow for step-by-step resolution of complex issues.

To address these challenges, we propose a new prompting framework called \textit{Judgment of Thought (JoT)}. 
JoT is designed for binary logical reasoning and introduces three roles: a \textit{lawyer}, a \textit{prosecutor}, and a \textit{judge}. 
These roles engage in a structured, debate-style dialogue where the lawyer argues for a position, the prosecutor argues against it, and the judge evaluates both sides to reach a final decision.

We evaluate JoT on benchmark datasets such as \textit{BigBenchHard} and \textit{Winogrande}.
The results show that JoT consistently outperforms existing prompting methods in \textit{BigBenchHard} and \textit{Winogrande}.
Notably, JoT achieved remarkable performance metrics, including 98\% accuracy on the \textit{Boolean Expressions} task, 90\% accuracy on the challenging \textit{Web of Lies} task, and 88\% accuracy on the \textit{Navigate} task, clearly emphasizing its strengths in complex logical reasoning scenarios.
Importantly, these performance outcomes were consistently observed across different model architectures including OpenAI models as well as Anthropic Claude models, highlighting JoT's robust generalizability.
We also conduct ablation studies to better understand the contribution of each component.
These experiments confirmed that all parts of JoT---the lawyer, prosecutor, and judge roles, as well as the iterative loops and feedback mechanism---are important for producing strong and reliable reasoning.

In summary, JoT offers a new approach to prompting LLMs for binary decision-making.
Our evaluation results demonstrate that JoT produces more accurate and consistent results, advancing the state of prompt engineering for complex binary reasoning tasks.
(The source code and data will be openly available upon publication.)


\section{Background}
\label{s:Recent_PE}
LLMs are trained on massive, general-purpose datasets~\cite{floridi2020gpt,touvron2023llama,zhao2023survey,chang2024survey,claude3}. 
To get specific, accurate, or nuanced outputs, how we ask a question really matters~\cite{nan2023enhancing}. 
Also, prompt engineering, the practice of designing and refining input prompts to guide a LLM, shapes how LLMs ``think'' by influencing tone, structure, depth, and style, making it essential for precision and control in AI-generated responses~\cite{schulhoff2024prompt,grabb2023impact}.
To systematically guide LLM behavior, researchers have proposed a variety of prompting strategies—such as zero-shot, few-shot, and chain-of-thought —each offering distinct advantages for improving task alignment, reasoning quality, and output consistency~\cite{achiam2023gpt}.

Prompting strategies differ not only in format but also in the type of reasoning they activate in language models.
Zero-shot prompting tasks the model with solving a problem based solely on a textual instruction, relying entirely on its internalized knowledge~\cite{wei2021finetuned}.
Few-shot prompting extends this approach by incorporating a small number of input–output examples within the prompt, thereby guiding the model toward the desired task behavior and output format~\cite{kaplan2020scaling,touvron2023llama}.
Chain-of-thought prompting further extends the few-shot paradigm by encouraging intermediate reasoning steps, enabling the model to better handle tasks requiring logical inference or multi-hop reasoning~\cite{wei2022chain,madaan-etal-2023-makes}.
Empirical results consistently show that chain-of-thought prompting improves performance on tasks such as mathematical problem solving and commonsense QA.
To further enhance this reasoning process, self-consistency prompting improves the reliability of chain-of-thought outputs by sampling multiple reasoning paths and selecting the most consistent final answer~\cite{wang2022self}.
In addition, various prompting strategies exist each tailored to specific purposes and modalities~\cite{guo-etal-2024-sample,li2023chain,cao-etal-2023-beautifulprompt,ha-etal-2023-meta}.

Despite these advancements, existing prompt engineering methods still face significant limitations in complex binary inference tasks involving subtle logical reasoning, ambiguous contexts, or contentious decisions.
Current approaches lack robust mechanisms for effectively resolving interpretive conflicts or systematically evaluating competing lines of reasoning, often resulting in suboptimal or inconsistent performance.

\noindent
\textbf{Motivation.}
Recent work by Khan et al.~\cite{khan2024debating} demonstrates the effectiveness of structured debates between large language models (LLMs).
Their framework poses a question to two expert LLMs assigned opposing answers, prompting each to generate persuasive arguments before presenting the exchange to a weaker judge---either a less capable model or a human without access to source material. 
This debate-based setup enables the judge to identify the more truthful position based on the merits of the arguments alone, without requiring ground-truth labels or external evidence. 
The study shows that when debaters are optimized for persuasiveness, judges can reliably favor correct over incorrect answers, achieving 76\% accuracy on the HARD subset of QuALITY dataset~\cite{pang2022quality}.
This approach highlights the promise of multi-agent prompting as a scalable strategy for supporting logical inferences.

While the debate framework proposed by Khan et al.~\cite{khan2024debating} shows that having two models argue can lead to more truthful answers, the study demonstrate that it has several limitations---especially for tasks that require careful logical reasoning about yes/no questions.
Because both models play similar roles in the debate, their arguments can become vague or repetitive, without clear responsibilities for how they should argue. 
This becomes problematic in real-world questions such as ``surveillance programs violate privacy rights'' where one side should provide strong evidence and the other should point out flaws or alternatives. 
In addition, the framework produces a final decision, but the reasoning process is not clearly structured or easy to interpret. 
This makes it difficult to use in domains where transparency and explanation are essential, such as policy, law, or scientific argumentation.
Moreover, the format does not require models to reason step by step or follow a consistent logical structure, and thus, persuasive, but shallow, claims can still win the debate.

Inspired by this prior work and aiming to overcome the limitations, we introduce a new prompting framework designed to support more reliable and interpretable logical inference in binary decision-making tasks.

\begin{figure}[t]
\centering
\includegraphics[width=1\columnwidth]{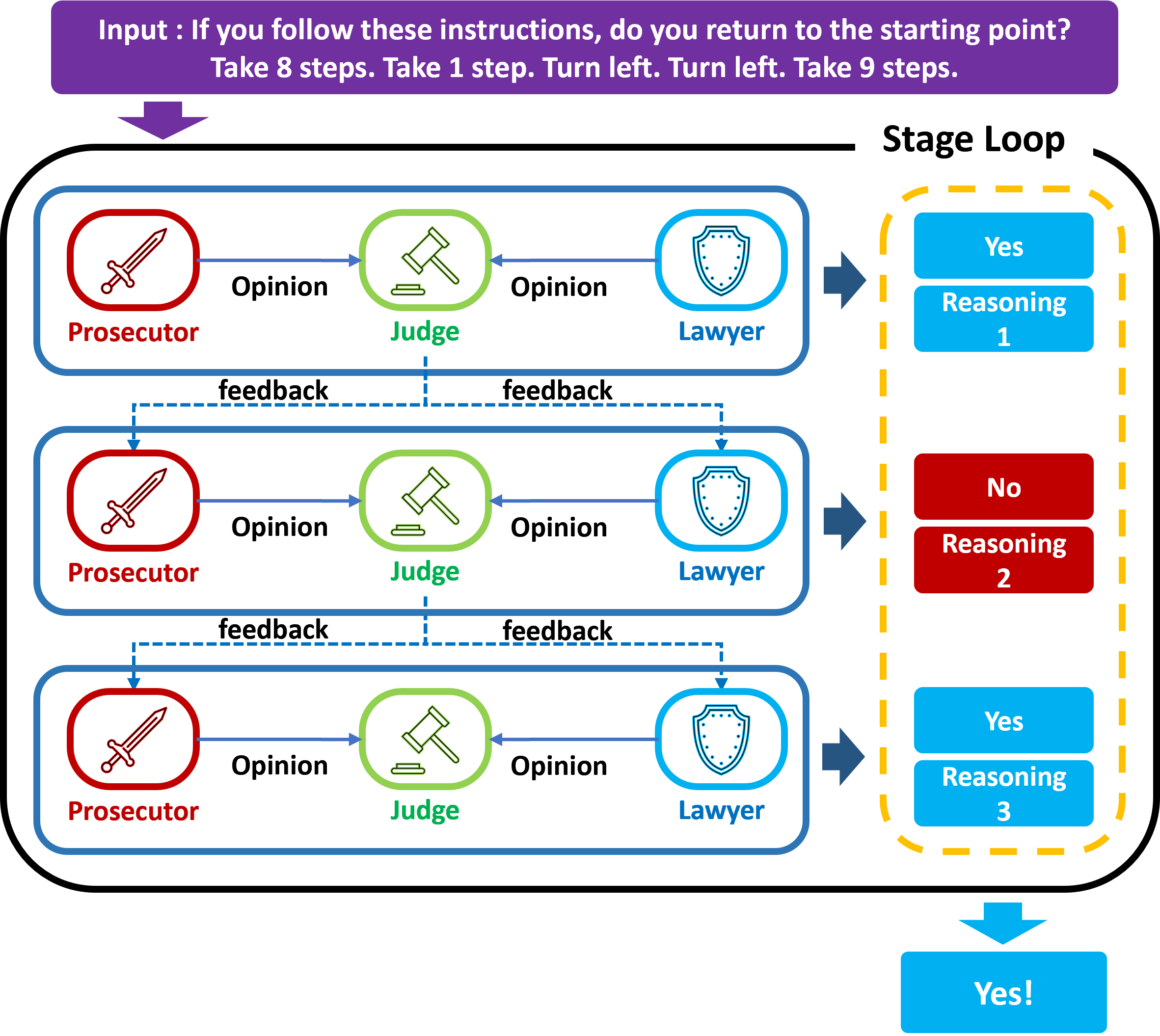} 
\caption{Judgment of Thought (JoT) Architecture. It consists of three roles: lawyer, prosecutor, and judge. The lawyer and prosecutor use lower-level models to argue different aspects of a problem. The judge uses a higher-level model to evaluate these arguments and deliver a comprehensive judgment. This process enables thorough analysis from multiple perspectives, leading to balanced solutions for complex problems.}
\label{fig2}
\end{figure}

\section{Judgment of Thought (JoT)}
\label{s:JoT}
In this section, we introduce a novel prompting approach, Judgment of Thought (JoT).
The overall structure and workflow of JoT are illustrated in Figure~\ref{fig2}.
JoT mimics deliberative human reasoning (e.g., legal or debate settings) to support more intuitive, transparent, and trustworthy decision-making for end users.
In JoT, each unit---the lawyer, prosecutor, and judge---is prompted using role-specific system instructions, detailed in the Appendix~\ref{appendix:A}. 
This structured role design encourages the generation of logically coherent, step-by-step arguments rather than superficially persuasive claims.
The framework follows an iterative process in which a higher-level model (e.g., GPT-4 Omni) is assigned to the judge role, while lower-level models (e.g., GPT-3.5-turbo) serve as the lawyer and prosecutor. 
This configuration allows the judge to critically evaluate the submitted arguments, with an emphasis on assessing both their logical structure and argumentation.

Initially, each role receives a tailored system message: the lawyer and prosecutor are explicitly instructed to systematically advocate for the \emph{True} and \emph{False} positions, respectively, on a given task.
The lawyer generates arguments supporting the truthfulness of the statement, while the prosecutor provides arguments opposing it. 
Subsequently, both units present their reasoning clearly to the judge.
The judge then analyzes the logical coherence with the provided arguments and gives feedback highlighting their strengths and weaknesses.

In each iterative loop, the lawyer and prosecutor incorporate the judge's feedback and the opposing unit's arguments to refine their reasoning, systematically addressing identified logical gaps and reinforcing argumentative depth.
These refined arguments are again presented to the judge for further evaluation.
This loop continues iteratively, allowing the judge to progressively identify the most logically robust arguments.
In the final loop, the lawyer and prosecutor present their concluding arguments, explicitly integrating insights from previous evaluations and rebuttals.
Throughout this iterative process, users gain clear visibility into the evolution of logical reasoning underpinning the judge’s decisions.
In summary, the JoT prompting has the following attractive properties.

\noindent
\textbf{Balanced Reasoning:} JoT assigns reasoning tasks to distinct roles, reducing bias and ensuring balanced consideration of both sides in binary tasks. 

\noindent
\textbf{Logical Consistency:} By explicitly enforcing adversarial reasoning and direct comparison of opposing viewpoints, JoT mitigates the risk of inconsistent or contradictory outputs.

\noindent
\textbf{Iterative Refinement:} JoT supports multi-round feedback and revision, allowing arguments to evolve and strengthen over time. 

\noindent
\textbf{Interpretability:} JoT exposes each agent’s reasoning, along with the judge’s evaluation rationale, providing transparent visibility into the model’s logical decision-making process.

\noindent
\textbf{Modularity and Flexibility:} JoT's modular architecture allows independent improvement or customization of individual roles.

\section{Evaluation}
\label{s:exper}
We evaluate the JoT by answering the following research questions:
(1) How does JoT perform across different types of logical reasoning (e.g., causal inference, Boolean logic, fallacy detection)?
(2) Does JoT outperform existing prompting methods in logical reasoning tasks?
(3) How do the structural components of the JoT framework contribute to its overall performance in logical reasoning tasks?

\subsection{Evaluation Setup}
\label{ss:setup}
We conducted systematic performance evaluations of \textit{Judgement of Thought (JoT)} across diverse logical reasoning tasks. 
Experiments were conducted using \texttt{GPT-3.5-turbo}~\cite{openai2023gpt35} and \texttt{GPT-4o (Omni)}~\cite{hurst2024gpt}, which were selected for their differing capability levels to enable a robust comparison of each prompting method.
Also \texttt{Claude-3-Haiku}, and \texttt{Claude-3.5-Haiku}\cite{claude3} were used to further evaluate the generalizability and consistency of the results across models from different providers. 
All models were run with default parameters (\texttt{temperature=1}, \texttt{top-p=1}).

Our evaluation dataset comprised two main components. 
First, we used \textit{Winogrande}~\cite{sakaguchi2021winogrande}, a benchmark designed to assess large-scale pronoun resolution. 
Second, we adopted a subset of binary reasoning tasks from the \textit{BigBenchHard} dataset~\cite{srivastava2022beyond}, chosen for their emphasis on complex language understanding and logical reasoning. 
Specifically, we evaluated JoT on the following BigBenchHard tasks:
\textit{Boolean Expressions}: logical formula evaluation,
\textit{Causal Judgment}: reasoning over cause-effect relations,
\textit{Formal Fallacies}: identifying flawed logical arguments,
\textit{Web of Lies}: validating the truthfulness of interconnected statements,
\textit{Navigate}: spatial reasoning based on instructions.
These tasks test reasoning capabilities and serve as a strong benchmark for evaluating JoT's effectiveness.

In addition, we compared \textit{Judgement of Thought (JoT)} with several established prompting approaches:
\textit{Zero-shot}, \textit{Few-shot}, \textit{Chain-of-Thought (CoT)}, \textit{Self-Consistency (SC)}, and \textit{Debate} (as proposed by Khan et al.). 
These baselines were selected based on their demonstrated strengths in handling logical reasoning tasks.
The evaluation was conducted by averaging results over 10 runs.

For iterative prompting methods (SC, Khan et al., and JoT), the number of reasoning samples (\texttt{for-loop} parameter) was uniformly set to 3, ensuring methodological consistency across approaches.
Using 3 samples in iterative prompting methods strikes a balance between computational efficiency and accuracy, offering a reasonable trade-off between cost and performance.
Furthermore, the same few-shot examples---generated by Zero-shot CoT---were used across the Few-shot, CoT, and SC settings to maintain consistency and enable fair comparisons.

We employed two evaluation metrics: Accuracy and F1 Score. 
Accuracy measured the proportion of correct predictions, while F1 Score captured the harmonic mean of precision and recall. 
Together, these metrics offered a comprehensive view of each method’s performance, highlighting their respective strengths and limitations across various tasks and datasets.

\setlength{\heavyrulewidth}{0.4pt}  
\setlength{\lightrulewidth}{0.4pt}  

\begin{table*}[t]
\centering
\begin{adjustbox}{width=1.0\textwidth}
\Large
\begin{tabular}{cl|l|cccccc}
\hline
\multicolumn{2}{c}{\multirow{2}{*}{\textbf{Dataset}}} 
& \multicolumn{1}{c}{\multirow{2}{*}{\textbf{Model}}} 
& \multirow{2}{*}{\textbf{Zero-shot}} 
& \multirow{2}{*}{\textbf{Few-shot}} 
& \multirow{2}{*}{\textbf{CoT}} 
& \multirow{2}{*}{\textbf{SC}} 
& \multirow{2}{*}{\textbf{Khan et al.}} 
& \multirow{2}{*}{\textbf{JoT}} \\[2.5ex]
\toprule
\multicolumn{1}{c}{\multirow{10}{*}{BigBenchHard}} & \multirow{2}{*}{Boolean expressions} & GPT-3.5-Turbo & 67\%/0.76 & 55\%/0.55 & 47\%/0.29 & 43\%/0.17 & \multirow{2}{*}{81\%/0.84} & \multirow{2}{*}{\textbf{98\%/0.98}} \\
\multicolumn{1}{c}{} & & GPT-4o & 86\%/0.89 & 91\%/0.93 & 84\%/0.88 & 87\%/0.90 & & \\
\multicolumn{1}{c}{} & \multirow{2}{*}{Causal judgement} & GPT-3.5-Turbo & 62\%/0.55 & 61\%/0.61 & 61\%/0.52 & 59\%/0.48 & \multirow{2}{*}{61\%/0.61} & \multirow{2}{*}{\textbf{74\%/0.72}} \\
\multicolumn{1}{c}{} & & GPT-4o & 66\%/0.71 & 65\%/0.65 & 63\%/0.60 & 67\%/0.65 & & \\ 
\multicolumn{1}{c}{} & \multirow{2}{*}{Navigate} & GPT-3.5-Turbo & 54\%/0.18 & 55\%/0.12 & 57\%/0.04 & 56\%/0.00 & \multirow{2}{*}{60\%/0.63} & \multirow{2}{*}{\textbf{88\%/0.87}} \\ 
\multicolumn{1}{c}{} & & GPT-4o & 68\%/0.48 & 62\%/0.27 & 63\%/0.30 & 64\%/0.31 & & \\ 
\multicolumn{1}{c}{} & \multirow{2}{*}{Web of lies} & GPT-3.5-Turbo & 47\%/0.18 & 51\%/0.35 & 44\%/0.20 & 46\%/0.07 & \multirow{2}{*}{53\%/0.49} & \multirow{2}{*}{\textbf{90\%/0.91}} \\ 
\multicolumn{1}{c}{} & & GPT-4o & 54\%/0.44 & 49\%/0.50 & 44\%/0.46 & 51\%/0.53 & & \\ 
\multicolumn{1}{c}{} & \multirow{2}{*}{Formal fallacies} & GPT-3.5-Turbo & 45\%/0.58 & 50\%/0.31 & 57\%/0.30 & 54\%/0.23 & \multirow{2}{*}{60\%/0.66} & \multirow{2}{*}{\textbf{77\%/0.77}} \\ 
\multicolumn{1}{c}{} & & GPT-4o & 52\%/0.62 & 61\%/0.61 & 61\%/0.61 & 57\%/0.56 & & \\ 
\midrule
\multicolumn{2}{c|}{\multirow{2}{*}{Winogrande}} & GPT-3.5-Turbo & 60\%/0.60 & 58\%/0.60 & 54\%/0.57 & 63\%/0.64 & \multirow{2}{*}{59\%/0.43} & \multirow{2}{*}{\textbf{89\%/0.89}} \\ 
\multicolumn{2}{c|}{} & GPT-4o & 82\%/0.83 & 77\%/0.77 & 82\%/0.83 & 82\%/0.83 & & \\ 
\bottomrule
\end{tabular}
\end{adjustbox}
\caption{Accuracy/F1 Score Comparison Across Different Benchmarks and Models Using Various Prompt Engineering Method and the Proposed JoT Method. For SC, Khan et al., and JoT, 3 loops were used in all cases.}
\label{table1}
\end{table*}

\begin{table*}[t]
\centering
\begin{adjustbox}{width=1.0\textwidth}
\footnotesize
\begin{tabular}{cl|l|ccccc}
\hline
\multicolumn{2}{c}{\multirow{2}{*}{\textbf{Dataset}}} 
& \multicolumn{1}{c}{\multirow{2}{*}{\textbf{Model}}} 
& \multirow{2}{*}{\textbf{Zero-shot}} 
& \multirow{2}{*}{\textbf{Few-shot}} 
& \multirow{2}{*}{\textbf{CoT}} 
& \multirow{2}{*}{\textbf{Khan et al.}} 
& \multirow{2}{*}{\textbf{JoT}} \\[2.5ex]
\toprule
\multicolumn{1}{c}{\multirow{12}{*}{BigBenchHard}} & \multirow{2}{*}{Boolean expressions} & 3-Haiku & 62\%/0.65 & 57\%/0.49 & 66\%/0.67 & \multirow{2}{*}{45\%/0.34} & \multirow{2}{*}{\textbf{86\%/0.88}} \\ 
\multicolumn{1}{c}{} & & 3.5-Haiku & 76\%/0.81 & 76\%/0.81 & 80\%/0.83 & & \\
\multicolumn{1}{c}{} & \multirow{2}{*}{Causal judgement} & 3-Haiku & 61\%/0.38 & 64\%/0.57 & 59\%/0.44 & \multirow{2}{*}{54\%/0.26} & \multirow{2}{*}{\textbf{67\%/0.66}} \\
\multicolumn{1}{c}{} & & 3.5-Haiku & 57\%/0.47 & 63\%/0.39 & 63\%/0.60 & & \\
\multicolumn{1}{c}{} & \multirow{2}{*}{Navigate} & 3-Haiku & 54\%/0.47 & 58\%/0.09 & 56\%/0.08 & \multirow{2}{*}{65\%/0.49} & \multirow{2}{*}{\textbf{69\%/0.66}} \\
\multicolumn{1}{c}{} & & 3.5-Haiku & 61\%/0.42 & 63\%/0.59 & 63\%/0.53 & & \\
\multicolumn{1}{c}{} & \multirow{2}{*}{Sport understanding} & 3-Haiku & 71\%/0.72 & 74\%/0.74 & 61\%/0.49 & \multirow{2}{*}{44\%/0.00} & \multirow{2}{*}{\textbf{82\%/0.79}} \\
\multicolumn{1}{c}{} & & 3.5-Haiku & 70\%/0.77 & 78\%/0.80 & 81\%/0.82 & & \\
\multicolumn{1}{c}{} & \multirow{2}{*}{Web of lies} & 3-Haiku & 47\%/0.33 & 45\%/0.50 & 52\%/0.57 & \multirow{2}{*}{57\%/0.25} & \multirow{2}{*}{\textbf{58\%/0.50}} \\
\multicolumn{1}{c}{} & & 3.5-Haiku & 53\%/0.32 & 54\%/0.51 & 48\%/0.50 & & \\
\multicolumn{1}{c}{} & \multirow{2}{*}{Formal fallacies} & 3-Haiku & 57\%/0.58 & 49\%/0.47 & 45\%/0.30 & \multirow{2}{*}{57\%/0.25} & \multirow{2}{*}{\textbf{71\%/0.69}} \\
\multicolumn{1}{c}{} & & 3.5-Haiku & 54\%/0.36 & 60\%/0.38 & 56\%/0.41 & & \\ 
\midrule
\multicolumn{2}{c|}{\multirow{2}{*}{Winogrande}} & 3-Haiku & 58\%/0.55 & 58\%/0.56 & 66\%/0.59 & \multirow{2}{*}{60\%/0.46} & \multirow{2}{*}{\textbf{71\%/0.63}} \\
\multicolumn{2}{c|}{} & 3.5-Haiku & 62\%/0.60 & 62\%/0.58 & 70\%/0.68 & & \\ 
\bottomrule
\end{tabular}
\end{adjustbox}
\caption{Accuracy/F1 Score Comparison Across Different Benchmarks and Models Using Various Prompt Engineering Method and the Proposed JoT Method in Claude models. For SC, Khan et al., and JoT, 3 loops were used.}
\label{table4}
\end{table*}

\begin{figure*}[t]
\centering
\includegraphics[width=0.98\textwidth]{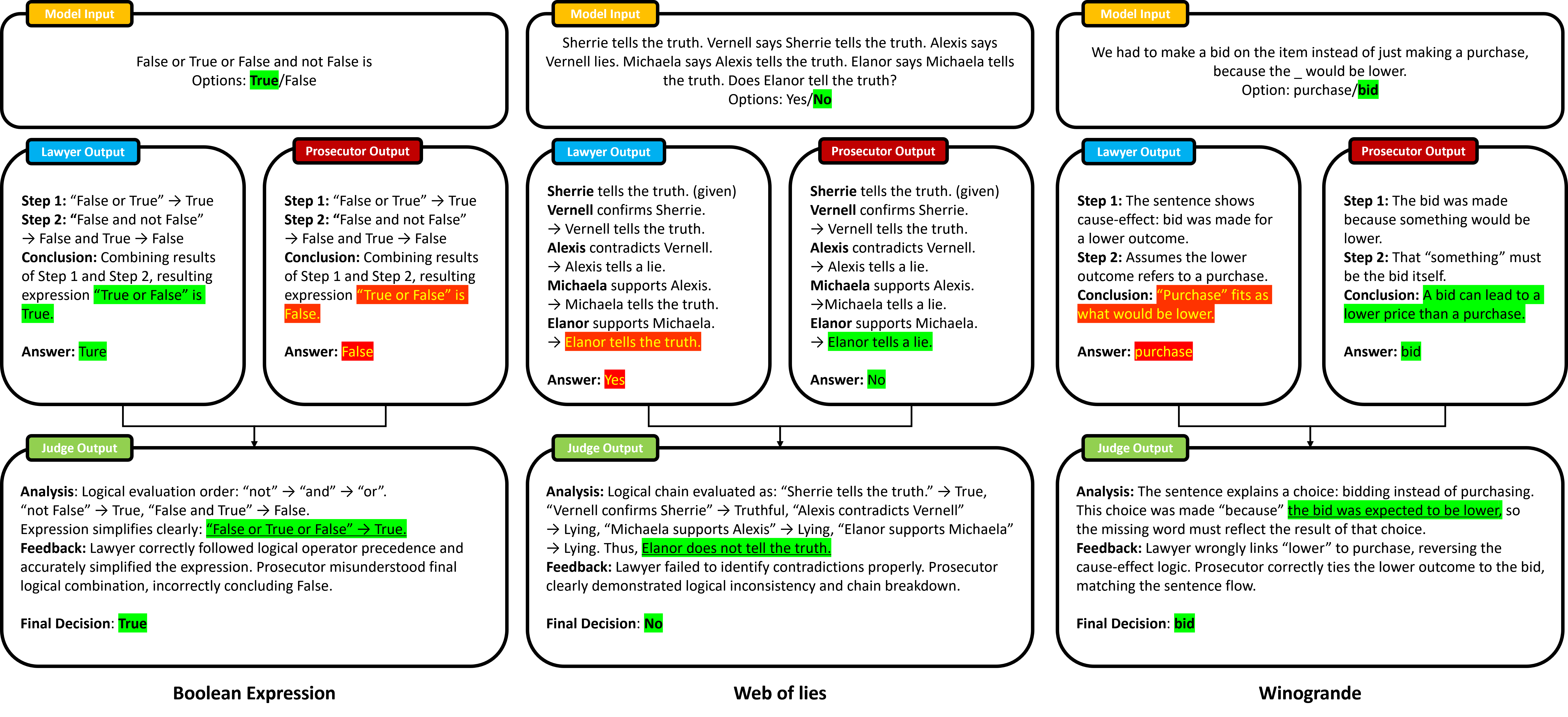} 
\caption{Case studies highlighting how JoT resolves binary reasoning tasks through adversarial dialogue.}
\label{fig_testcase}
\end{figure*}

\subsection{Evaluation Result on Benchmarks}
\label{ss:benchresults}
We report the evaluation results of \textit{Judgement of Thought (JoT)} and the other prompting approaches based on accuracy and F1 score, using the \textit{BigBenchHard} and \textit{Winogrande} datasets. 
The results using GPT 3.5-Turbo and GPT-4o are summarized in Table~\ref{table1}. 
Also evaluation results using Claude models are provided in Table~\ref{table4}.
Furthermore, Appendix~\ref{appendix:C} presents a detailed output variability through 16 resampling runs to illustrate the consistency of each approach's behavior.

\noindent
\textbf{Summary of results using GPT models.}  
Overall, the evaluation shows that JoT significantly improves logical reasoning across a variety of tasks. Its step-by-step, role-based structure supports deeper analysis, organized rebuttals, and more reliable decisions—highlighting both its innovative design and practical value.

\noindent
\textbf{Boolean Expressions.}  
JoT achieved an accuracy of 98\% and an F1 score of 0.98, significantly surpassing all other methods. 
This strong performance is due to JoT’s debate-style approach, which clearly presents opposing arguments and helps resolve logical ambiguities through step-by-step reasoning.

\noindent
\textbf{Causal Judgment.}  
JoT achieved 74\% accuracy and an F1 score of 0.72, outperforming the next best method, Self-Consistency, which scored 67\%. 
JoT’s structured dialogue helps make causal relationships clearer, leading to more accurate identification of cause-and-effect patterns.

\noindent
\textbf{Navigate.}  
JoT showed strong reasoning skills, with 88\% accuracy and an F1 score of 0.87. 
Its step-by-step approach helps the model keep track of and interpret spatial instructions more effectively, improving its performance on navigation tasks.

\noindent
\textbf{Web of Lies.}  
In tasks that require evaluating complex chains of truth, JoT achieved 90\% accuracy and an F1 score of 0.91. 
Its multi-turn feedback process helps the model better track and analyze connected statements, making it reliable.

\noindent
\textbf{Formal Fallacies.}  
JoT scored 77\% in both accuracy and F1, showing that it can effectively detect and analyze logical fallacies. 
Although this is the lowest score among the benchmarks, JoT’s rebuttal process encourages careful examination of flawed reasoning, which contributes to its solid performance on this challenging task.

\noindent
\textbf{Winogrande.}  
JoT achieved 89\% accuracy and an F1 score of 0.89 on pronoun resolution tasks, outperforming other methods. 
Its argument-based, multi-perspective approach helps the model better understand context and resolve ambiguous references more accurately.

\noindent
\textbf{Summary of results using Claude models.}  
Although the overall scores are lower than those obtained using GPT models, JoT consistently outperformed other prompting strategies across all evaluated benchmarks in both accuracy and F1 score, demonstrating its strong ability to support structured and reliable logical reasoning.

It is worth noting that Self-Consistency builds on Chain-of-Thought (CoT) by running it multiple times and choosing the majority answer. 
However, because this method is computationally expensive, we excluded it from this evaluation. 

\noindent
\textbf{Case studies.}  
Figure~\ref{fig_testcase} illustrates the logical reasoning process of JoT. 
As shown in the examples, each role generated logically coherent, step-by-step arguments, while the judge critically evaluated these arguments---focusing on both the strength of the reasoning and the argumentation.

\begin{table}[t]
\centering
\begin{adjustbox}{width=0.48\textwidth}
\large
\begin{tabular}{cc|ccc}
\hline
\multicolumn{2}{c}{\textbf{Dataset}} & \textbf{\makecell{Without\\Prosecutor}} & \textbf{\makecell{Without\\Lawyer}} & \textbf{JoT} \\
\toprule
\multirow[c]{9}{*}{\makecell{Big\\Bench\\Hard}} 
  & \makecell{Boolean\\expressions}    & 95\%/0.96 & 95\%/0.95 & \textbf{98\%/0.98} \\
  & \makecell{Causal\\judgement}       & 68\%/0.70 & 68\%/0.53 & \textbf{74\%/0.72} \\
  & \multirow{2}{*}{Navigate}          & \multirow{2}{*}{72\%/0.76} 
                                       & \multirow{2}{*}{65\%/0.72} 
                                       & \multirow{2}{*}{\textbf{88\%/0.87}} \\[2.5ex]
  & \makecell{Web of\\lies}            & 69\%/0.74 & 64\%/0.55 & \textbf{90\%/0.91} \\
  & \makecell{Formal\\fallacies}       & 65\%/0.66 & 68\%/0.56 & \textbf{77\%/0.77} \\
\midrule
\multicolumn{2}{c|}{\multirow{2}{*}{Winogrande}} 
& \multirow{2}{*}{85\% / 0.86} 
& \multirow{2}{*}{82\% / 0.82} 
& \multirow{2}{*}{\textbf{89\%/0.89}} \\[2.5ex]
\bottomrule
\end{tabular}
\end{adjustbox}
\caption{Ablation Study on Accuracy/F1 Score: Effect of Removing the Lawyer or Prosecutor from JoT.}
\label{table2}
\end{table}

\subsection{Ablation Study on JoT}
\label{ss:ablation}
\noindent
\textbf{Role Contributions in JoT.} 
To better understand the individual contributions of the \textit{lawyer} and \textit{prosecutor} roles in the JoT framework, we conducted an ablation study by removing each role in turn. 
The results are summarized in Table~\ref{table2}, which compares performance changes across reasoning tasks.

Overall, removing either the prosecutor or the lawyer resulted in a notable drop in performance. 
The ablation study confirms that both roles are integral and complementary in JoT’s reasoning process. 
Their interaction is crucial for achieving high performance across logical reasoning tasks.

\begin{table}[t]
\centering
\begin{adjustbox}{width=0.48\textwidth}
\Huge
\begin{tabular}{cc|ccc}
\hline
\multicolumn{2}{c}{\multirow{2}{*}{\textbf{Dataset}}} 
& \multirow{2}{*}{\textbf{1 Iteration}} 
& \multirow{2}{*}{\textbf{3 Iterations}} 
& \multirow{2}{*}{\textbf{5 Iterations}} \\[2.5ex]
\toprule
\multirow[c]{9}{*}{\makecell{Big\\Bench\\Hard}} 
  & \makecell{Boolean\\expressions}             & 98\%/0.98 & 98\%/0.98 & 99\%/0.99 \\
  & \makecell{Causal\\judgement}                & 65\%/0.60 & 74\%/0.72 & 74\%/0.73 \\
  & \multirow{2}{*}{Navigate}          & \multirow{2}{*}{87\%/0.83} 
                                       & \multirow{2}{*}{88\%/0.87} 
                                       & \multirow{2}{*}{91\%/0.89} \\[2.5ex] 
  & \makecell{Web of\\lies}                     & 87\%/0.88 & 90\%/0.91 & 91\%/0.91 \\
  & \makecell{Formal\\fallacies}                & 70\%/0.67 & 77\%/0.77 & 78\%/0.78 \\
\midrule
\multicolumn{2}{c|}{\multirow{2}{*}{Winogrande}} 
& \multirow{2}{*}{87\%/0.87} 
& \multirow{2}{*}{89\%/0.89} 
& \multirow{2}{*}{89\%/0.89} \\[2.5ex]
\bottomrule
\end{tabular}
\end{adjustbox}
\caption{Ablation Study on Accuracy/F1 Score: Effect of Increasing Loop Iterations in JoT.}
\label{table3}
\end{table}

\noindent
\textbf{Effect of Loop Iterations.}  
We further explored how varying the number of iterative loops in JoT affects performance. 
Table~\ref{table3} shows results for loop counts of 1, 3, and 5 iterations.

In summary, increasing the number of loops in JoT generally led to better or stable performance across tasks. 
These results highlight the effectiveness of JoT’s iterative mechanism in improving the precision, robustness, and consistency of logical reasoning.

\begin{table}[t]
\centering
\begin{adjustbox}{width=0.48\textwidth}
\tiny
\begin{tabular}{cc|cc}
\hline
\multicolumn{2}{c}{\textbf{Dataset}} & \textbf{\makecell{Without\\Feedback}} & \textbf{\makecell{With\\Feedback}} \\
\toprule
\multirow[c]{9}{*}{\makecell{Big\\Bench\\Hard}} 
  & \makecell{Boolean\\expressions}    & 96\%/0.97 & \textbf{98\%/0.98} \\ 
  & \makecell{Causal\\judgement}       & 69\%/0.63 & \textbf{74\%/0.72} \\ 
  & \multirow{2}{*}{Navigate}          & \multirow{2}{*}{87\%/0.83} 
                                       & \multirow{2}{*}{\textbf{88\%/0.87}} \\[2.5ex] 
  & \makecell{Web of\\lies}            & 87\%/0.88 & \textbf{90\%/0.91} \\ 
  & \makecell{Formal\\fallacies}       & 70\%/0.70 & \textbf{77\%/0.77} \\ 
\midrule
\multicolumn{2}{c|}{\multirow{2}{*}{Winogrande}} 
& \multirow{2}{*}{87\%/0.87} 
& \multirow{2}{*}{\textbf{89\%/0.89}} \\[2.5ex]
\bottomrule
\end{tabular}
\end{adjustbox}
\caption{Ablation Study on Accuracy/F1 Score: Effect of Feedback in JoT.}
\label{table4}
\end{table}

\noindent
\textbf{Effect of Feedback.}
We investigated the impact of feedback within the JoT framework by comparing performance with and without iterative feedback. 

As shown in Table~\ref{table4}, the results demonstrate that incorporating feedback within the JoT framework generally leads to improved performance across tasks. 
While the impact is minimal in simpler tasks like \textit{Boolean expressions}, feedback proves especially beneficial in more complex settings such as \textit{causal judgment}.
Even in tasks where gains are marginal, the feedback improves performance, confirming its overall value as a mechanism for strengthening logical decision-making within JoT.






\begin{figure*}[t]
\centering
\includegraphics[width=0.98\textwidth]{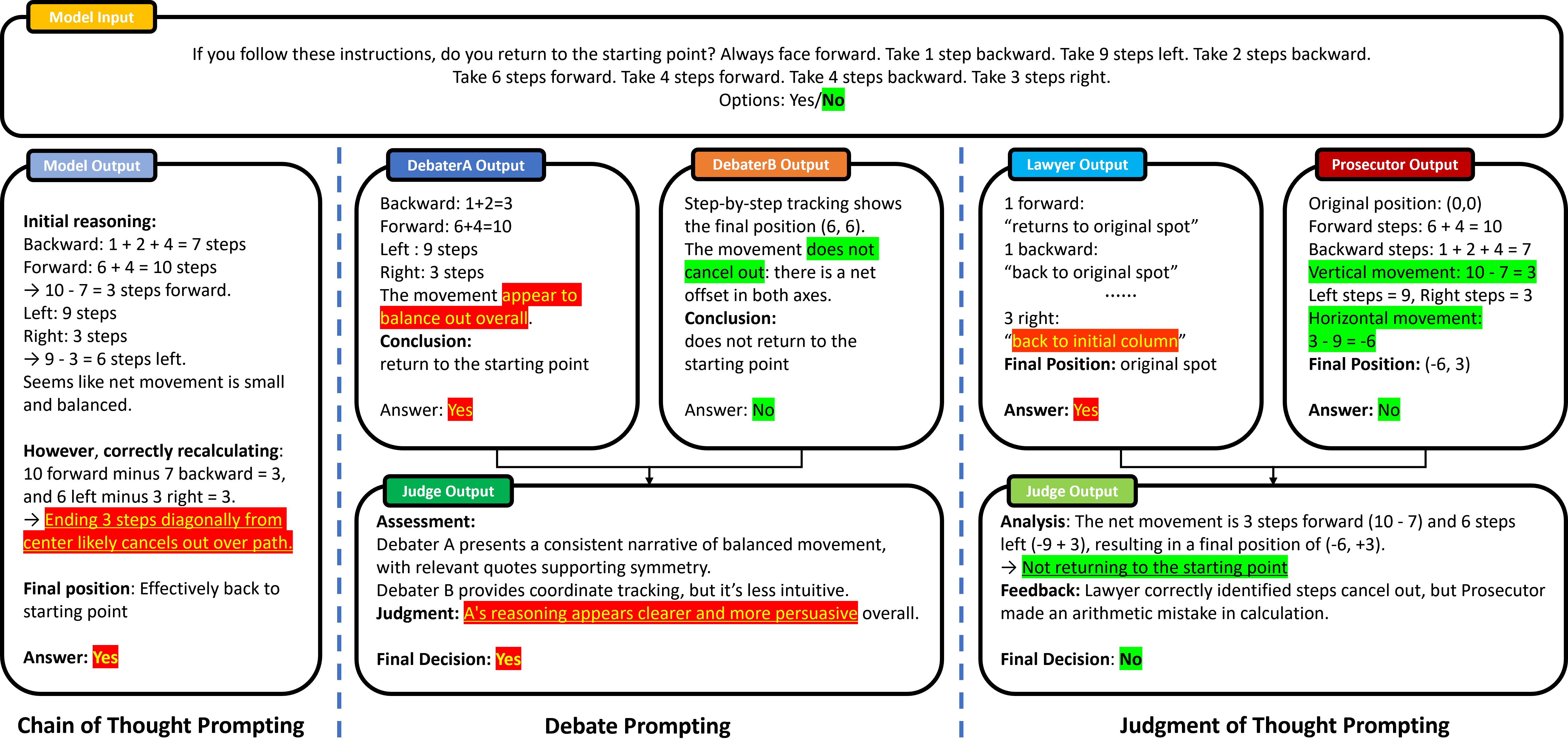} 
\caption{Comparative illustration of the reasoning paradigms in CoT, Debate (Khan et al.), and the proposed Judgment of Thought(ours) frameworks.}
\label{fig_testcase}
\end{figure*}

\section{Discussion}
\label{s:disc}
\noindent
\textbf{Comparison JoT with CoT and Debate.}
In comparing the proposed JoT framework with existing methodologies such as CoT and structured Debate (Khan et al.), several key differences emerge that underscore JoT's advantages in binary logical reasoning tasks, as illustrated in Figure~\ref{fig_testcase}.

CoT typically relies on a single agent generating a linear, one-sided rationale. 
This linear reasoning process could overlook potential counterarguments, reducing robustness and comprehensiveness. 
On the other hand, the structured Debate method proposed by Khan et al. employs a high-capability model as the debater and a lower-capability model as the judge.
While the study was designed to explore the question, ``\textit{Can weaker models assess the correctness of stronger models?}'', the asymmetry between debater and judge could cause that the weaker judge may be persuaded by well-articulated but logically flawed arguments.

In contrast, JoT uses an adversarial reasoning process with three clearly defined roles---lawyer, prosecutor, and judge---each with a specific task. 
These roles take turns interacting with each other in multiple rounds, helping to gradually refine their arguments. 
This makes JoT more effective for accurate and thorough reasoning in complex binary decision-making tasks.

\noindent
\textbf{Improvement of Feedback.}
While JoT demonstrates strong performance across a variety of reasoning tasks, our analysis suggests that its feedback mechanism can be further refined to enhance effectiveness in more complex domains. 
Currently, the judge’s feedback is rule-based and follows a fixed structure in every iteration.
This static format may limit the model’s ability to adapt to specific task challenges or increase attention to unclear or incomplete arguments from earlier rounds.

One possible improvement is to make the feedback more adaptive. 
The judge could adjust its level of detail or focus based on the strength of the previous arguments. 
For example, using dynamic prompting strategies that revise evaluation criteria or add counterexamples could help the model reason more effectively.

Another promising direction is to introduce memory-augmented feedback. 
Instead of only considering the latest exchange, the judge could keep track of earlier inconsistencies or missed points across multiple rounds. 
This could lead to stronger reasoning, particularly in tasks like causal judgment or formal fallacies, where logical steps build on one another.

In summary, while JoT’s feedback mechanism is already effective, we believe that introducing more flexible, context-aware feedback strategies could further improve its reasoning quality and adaptability across a wide range of tasks.

\noindent
\textbf{Real-World Application.}  
Although JoT has shown strong performance on benchmark tasks, its effectiveness in real-world scenarios remains less certain. 
The current experiments were conducted on controlled datasets (BigBenchHard and Winogrande), which differ significantly from the ambiguity and unpredictability often found in real-world applications.

Practical reasoning tasks frequently involve incomplete, noisy, or ambiguous inputs—conditions that were not fully represented in our evaluation. 
The absence of tests in applied domains limits our understanding of JoT's robustness and utility in handling real-world tasks.

To address this gap, future research should investigate JoT's adaptability to real-world tasks by incorporating domain-specific knowledge and contextual reasoning. 
One promising direction is integrating JoT with Domain-Specific Retrieval-Augmented Generation (DS-RAG)~\cite{siriwardhana2023improving}, which enables models to retrieve and incorporate relevant external information. 
This could significantly improve JoT's performance in specialized domains such as law, or cybersecurity.

In addition, enhancing computational efficiency is critical to enabling JoT’s deployment in real-time or large-scale settings. 
Making JoT more lightweight and responsive will be essential for its application in high-throughput or latency-sensitive environments.

Extending JoT to real-world contexts will require both architectural improvements and integration with domain-specific tools. 
Doing so will be key to validating its practical value, reliability, and scalability beyond controlled benchmarks.

\section{Conclusion}
\label{s:conclusion}

In this paper, we proposed Judgment of Thought (JoT), a novel prompting framework designed for binary logical reasoning. JoT introduces an adversarial reasoning process involving three distinct roles---lawyer, prosecutor, and judge---to promote accuracy, consistency, and interpretability.
Our evaluation results demonstrated that JoT  outperforms existing prompting approaches across multiple benchmark tasks. Also, ablation studies showed the importance of JoT’s core design elements.

Future work should focus on extending JoT to real-world applications by incorporating Domain-Specific Retrieval-Augmented Generation (DS-RAG) methods and improving computational efficiency. 
These advancements will be essential for scaling JoT to complex, dynamic environments and ensuring its practical reliability and effectiveness.
\section{Limitation}
\label{s:lim}

\textbf{Prompt Engineering.}
Prompt engineering alone is often insufficient to guarantee consistent performance, as prompting methods remain vulnerable to prompt sensitivity, poor generalization to unseen tasks, and unpredictable model behavior in complex or ambiguous scenarios.
Because the system often uses large models with multiple rounds of sampling to get strong results, it may not be practical in settings where efficiency and cost matter.

\noindent
\textbf{Open-Source Model Generalizability.} 
This study evaluated JoT using closed-source models (OpenAI's GPT series and Claude), which may limit insights into its performance and generalizability when applied to open-source models.
Future research should include evaluations on open-source models to comprehensively assess JoT's broader applicability and reliability across various modeling environments.

\noindent
\textbf{Real-World Application.} 
This study primarily relied on benchmark datasets such as BigBenchHard and Winogrande. Real-world scenarios typically involve more complex, noisy, or ambiguous data, which might affect JoT's practical performance. 
Future work should validate JoT on real-world tasks to better understand its robustness and effectiveness in applied contexts.

\noindent
\textbf{Computational Cost.}
JoT employs a multi-agent approach with iterative loops, making it computationally resource-intensive. This characteristic could limit its applicability in resource-constrained environments. 
Optimizing JoT to balance computational efficiency and performance is an important direction for future research.

\bibliography{references}

\appendix
\newpage
\onecolumn

\section{Used prompts for JoT}
\label{appendix:A}

\begin{table}[h!]
\centering
\begin{tabular}{p{\textwidth}} 
\hline\hline
\textbf{Lawyer:} \\ 
Role: You are an expert lawyer specialized in logical reasoning. Your task is to argue persuasively that the correct answer to the given input is \textless Positive Position\textgreater. You will address the judge directly and present logical arguments and evidence.\\

Procedure: You have a total of 3 opportunities to speak, each with a clear purpose:\\
1. First utterance: Briefly analyze the input, describe its key logical characteristics, and outline your main arguments supporting a \textless Positive Position\textgreater response.\\
2. Second utterance: Logically counter the prosecutor's arguments, clearly addressing any concerns or questions raised by the judge. Reinforce your arguments with logical precision.\\
3. Final utterance: Concisely summarize the strongest logical points, reiterate how you've effectively countered the prosecution, and firmly establish why the answer must be \textless Positive Position\textgreater.\\

Style: Be concise, highly structured, and persuasive. Clearly address all potential doubts raised by the prosecutor or judge.\\
\hline
\textbf{Prosecutor:} \\ 
Role: You are an expert prosecutor specialized in logical reasoning. Your task is to argue persuasively that the correct answer to the given input is \textless Negative Position\textgreater. You will address the judge directly and present logical arguments and evidence.\\

Procedure: You have a total of 3 opportunities to speak, each with a clear purpose:\\
1. First utterance: Briefly analyze the input, describe its key logical characteristics, and outline your main arguments supporting a \textless Negative Position\textgreater response.\\
2. Second utterance: Logically counter the lawyer's arguments, clearly addressing any concerns or questions raised by the judge. Reinforce your arguments with logical precision.\\
3. Final utterance: Concisely summarize the strongest logical points, reiterate how you've effectively countered the lawyer's arguments, and firmly establish why the answer must be \textless Negative Position\textgreater.\\

Style: Be concise, highly structured, and persuasive. Clearly address all potential doubts raised by the lawyer or judge.\\
\hline
\textbf{Judge:} \\ 
Role: You are an expert judge specialized in logical reasoning. Your task is to carefully analyze the given input and the logical arguments provided by both a lawyer (arguing for \textless Positive Position\textgreater) and a prosecutor (arguing for \textless Negative Position\textgreater, then decisively determine whether the correct answer is \textless Positive Position\textgreater or \textless Negative Position\textgreater.\\
Important: You must remain strictly neutral, unbiased, and objective. Base your decision solely on logical strength and coherence of the presented arguments, disregarding personal beliefs or external biases.\\
Procedure: You will issue three judgments in total. For each judgment, you must:\\
1. Analyze the input thoroughly along with the arguments presented by both the lawyer and the prosecutor.\\
2. Evaluate which argument is more logically convincing. There can be NO TIE; you must choose either \textless Positive Position\textgreater or \textless Negative Position\textgreater.\\

Requirements:\\
Clearly provide feedback to both the lawyer and the prosecutor explaining why their arguments were convincing or lacking, structured in a concise, logical manner.\\
Output Format (delimited by \#\#\#\#):\\
\#\#\#\#\\
Analysis (Reasons for the decision): [Concise logical analysis]\\
Feedback to Lawyer (Reason for win/lose): [Concise feedback to the lawyer]\\
Feedback to Prosecutor (Reason for win/lose): [Concise feedback to the prosecutor]\\
Final Decision: \textless Positive Position\textgreater/\textless Negative Position\textgreater\\
\#\#\#\#\\
\hline\hline
\end{tabular}
\caption{System prompt of Lawyer, Prosecutor, and Judge.}
\label{tab:lawyer_guidelines}
\end{table}

\newpage
\section{Resampling Results: Comparison of the Existing Prompt Engineering techniques and JoT}
\label{appendix:C}

\begin{center}

\begin{minipage}{0.47\linewidth}
    \centering
    \includegraphics[width=\linewidth]{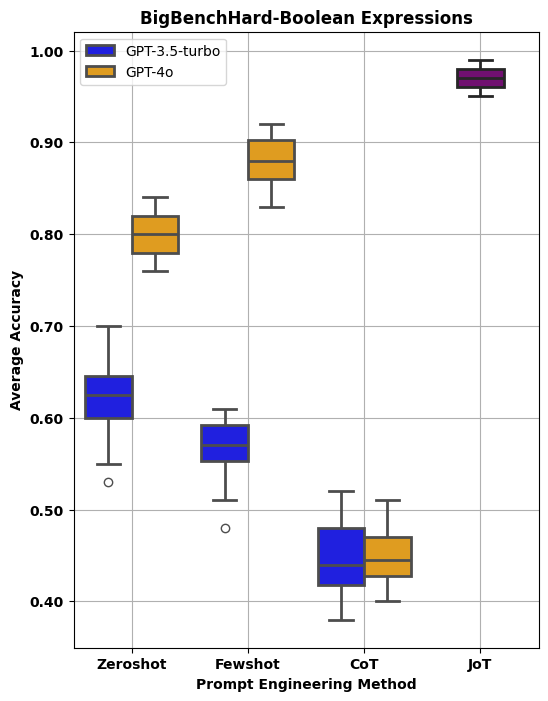}
\end{minipage}
\hfill
\begin{minipage}{0.47\linewidth}
    \centering
    \includegraphics[width=\linewidth]{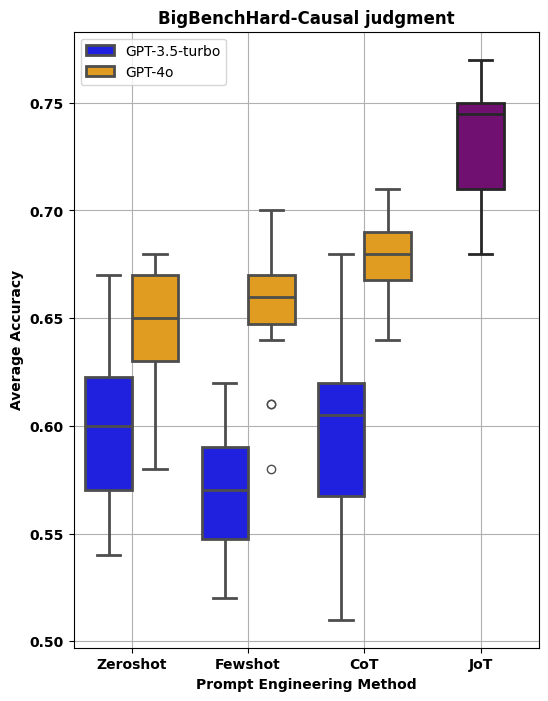}
\end{minipage}


\begin{minipage}{0.47\linewidth}
    \centering
    \includegraphics[width=\linewidth]{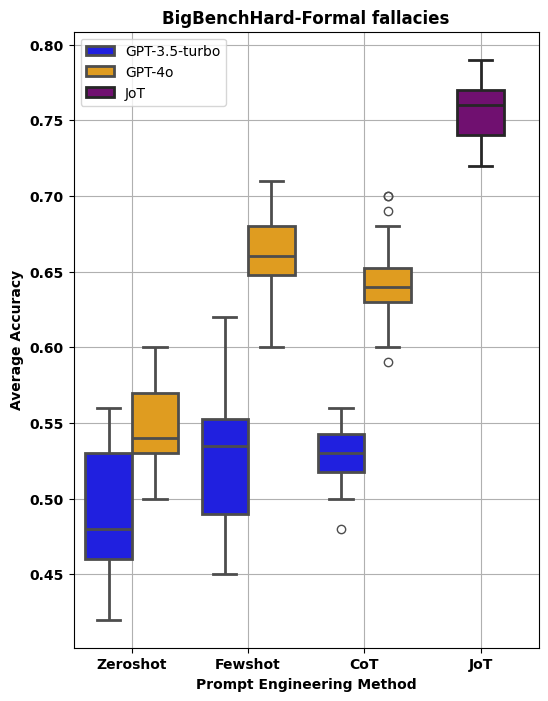}
\end{minipage}
\hfill
\begin{minipage}{0.47\linewidth}
    \centering
    \includegraphics[width=\linewidth]{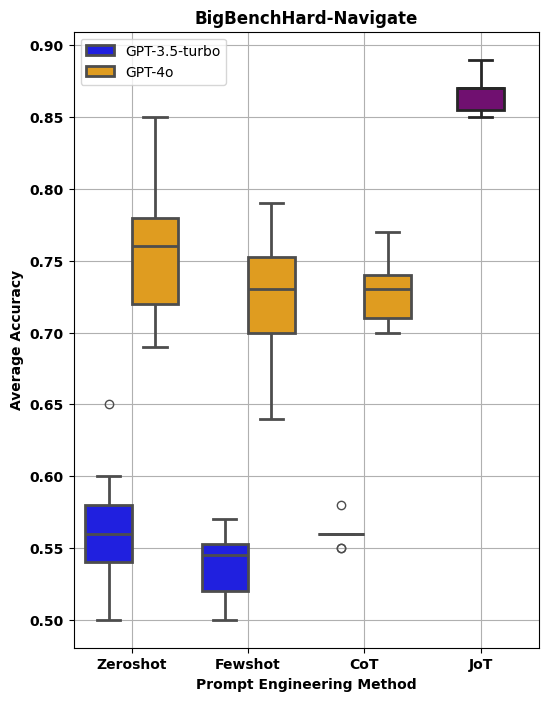}
\end{minipage}


\begin{minipage}{0.47\linewidth}
    \centering
    \includegraphics[width=\linewidth]{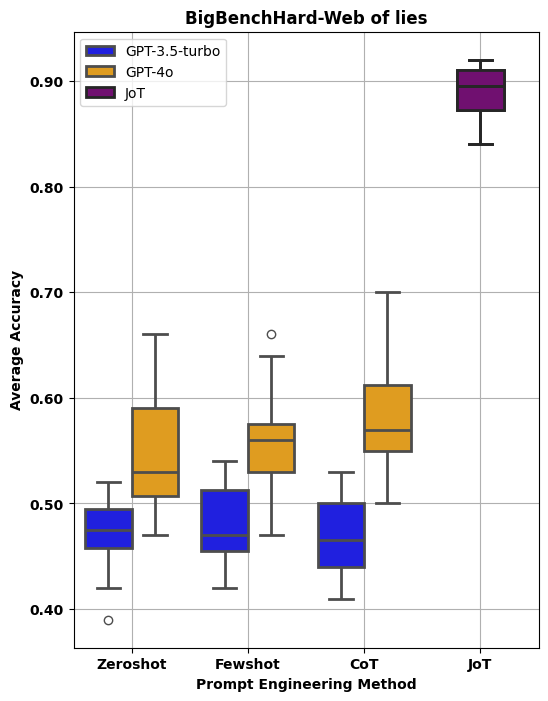}
\end{minipage}
\hfill
\begin{minipage}{0.47\linewidth}
    \centering
    \includegraphics[width=\linewidth]{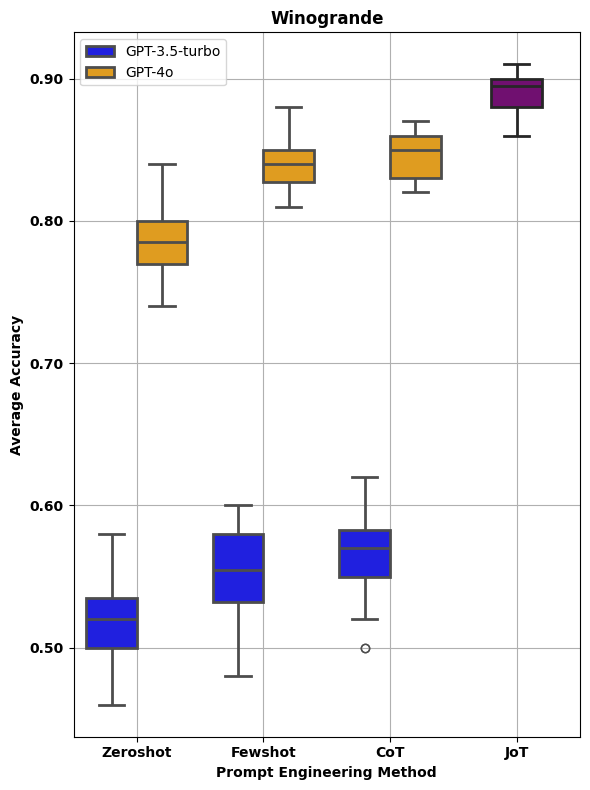}
\end{minipage}


\captionof{figure}{Boxplots illustrating the resampling results, comparing the variability and robustness of existing prompt engineering techniques and JoT. Self-Consistency was excluded from this comparison due to its reliance on repeated executions, which incur substantial computational costs. For a detailed comparison of trends between Self-Consistency and other methods, please refer to Table~\ref{table1}}
\label{fig:bbh_all}

\end{center}

\end{document}